\documentclass{article}

\usepackage{neurips_2025}

\usepackage[utf8]{inputenc} % allow utf-8 input
\usepackage[T1]{fontenc}    % use 8-bit T1 fonts
\usepackage{hyperref}       % hyperlinks
\usepackage{url}            % simple URL typesetting
\usepackage{booktabs}       % professional-quality tables
\usepackage{amsfonts}       % blackboard math symbols
\usepackage{nicefrac}       % compact symbols for 1/2, etc.
\usepackage{microtype}      % microtypography
\usepackage{xcolor}         % colors
\usepackage{graphicx}
\usepackage{multirow}
\usepackage{booktabs}
\usepackage{subcaption}
\usepackage{xcolor}         % colors
\usepackage{graphicx}
\usepackage{algorithm}
\usepackage{algpseudocode}
\usepackage{amsmath}

\title{Continuous Self-Improvement of Large Language Models by Test-time Training with Verifier-Driven Sample Selection}
\author{%
  Mohammad Mahdi Moradi\\
  Department of Computer Science, Concordia University\\
  Ascend Team, Huawei Technologies\\
  \texttt{mohammad.mahdi.moradi@h-partners.com} \\
  \And
  Hossam Amer\\
  Ascend Team \\
  Toronto Research Center \\
  Huawei Technologies \\
  \AND
  Sudhir  Mudur \\
  Department of Computer Science \\
  Concordia University \\
  \texttt{mudur@cs.concordia.ca} \\
  \And
  Weiwei Zhang\\
  Ascend Team \\
  Toronto Research Center \\
  Huawei Technologies \\
  \And
  Yang Liu\\
  Ascend Team \\
  Toronto Research Center \\
  Huawei Technologies \\
  \And
  Walid Ahmed\\
  Ascend Team \\
  Toronto Research Center \\
  Huawei Technologies \\
}

\begin{document}

\maketitle

\begin{abstract} \label{sec:abstract}
  Learning to adapt pretrained language models to unlabeled, out‑of‑distribution data is a critical challenge, as models often falter on structurally novel reasoning tasks even while excelling within their training distribution. We introduce a new framework called VDS-TTT - Verifier‑Driven Sample Selection for Test‑Time Training to efficiently address this. We use a learned verifier to score a pool of generated responses and select only from high ranking pseudo‑labeled examples for fine-tuned adaptation. Specifically, for each input query our LLM generates $N$ candidate answers; the verifier assigns a reliability score to each, and the response with the highest confidence and above a fixed threshold is paired with its query for test‑time training. We fine‑tune only low‑rank LoRA adapter parameters, ensuring adaptation efficiency and fast convergence. Our proposed self-supervised framework is the first to synthesize verifier driven test-time training data for continuous self-improvement of the model. Experiments across three diverse benchmarks and three state‑of‑the‑art LLMs demonstrate that VDS‑TTT yields up to a 32.29\% relative improvement over the base model and a 6.66\% gain compared to verifier-based methods without test‑time training, highlighting its effectiveness and efficiency for on‑the‑fly large language model adaptation.
\end{abstract}

\section{Introduction} \label{sec:Introduction}
Large-scale language models (LLMs) are typically developed under a fixed training-testing paradigm, where model parameters are learned once and remain unchanged during test time. However, a long-standing idea in machine learning advocates for test-time adaptation, i.e., updating the model using information specific to each test instance. This highlights the distinction between inductive learning, which aims to generalize from training data to unseen examples, and transductive learning, which enables models to dynamically adapt at test time. By embracing test-time updates, LLMs can better handle domain shifts and nonstationary environments ( \cite{xiao2024beyond}).

Test‑time training (TTT) is a paradigm that revisits this idea, updating model parameters at test time based on each incoming test instance or batch, thereby performing on‑the‑fly transductive learning. Transductive or local learning methods, such as locally weighted regression and local fine‑tuning—have been studied since the early days of machine learning (\cite{bottou1992local, cleveland1979robust, cleveland1988locally}), demonstrating the potential of instance‑specific adaptation. TTT differs from continual pretraining. TTT adapts a pre-trained model during test time to each test instance using unsupervised objectives, requiring minimal or no labeled data and achieving rapid task-specific tuning.  In contrast, Continual Pretraining (CPT) (\cite{wu2024continual}) continuously updates a model on large, streaming datasets, supervised or unsupervised, aiming to gradually enrich its generalization capabilities across evolving domains.
%Test‑time training (TTT) is a paradigm that revisits this idea, updating model parameters at inference based on each incoming test instance, thereby performing on‑the‑fly transductive learning. Transductive or local learning methods—such as locally weighted regression and local fine‑tuning—have been studied since the early days of machine learning \cite{bottou1992local, cleveland1979robust, cleveland1988locally}, demonstrating the potential of instance‑specific adaptation. Relatively more recent line of work treats an RNN’s hidden state as model parameters and the state‑update equation is reinterpreted as solving a recall‑based regression objective \cite{sun2407learning, behrouz2024titans, wang2025test, karami2025lattice}.  Since the hidden state is updated by training even on test sequences, this is considered as TTT, of course with significant storage overhead to maintain the hidden state. With the advent of large pre‑trained LLMs that offer rich, transferable representations, a different approach is pursued, namely Test Time Scaling. The idea is to use extra compute during inference to improve performance with or without training, especially for complex reasoning tasks\cite{zhang2025whathowwherewell}. Rapid task‑specific calibration at inference time is used to improve performance under distributional shifts. Our proposed approach presented in this paper is based on this paradigm. 

A fundamental challenge in TTT is adapting to unseen test distributions without access to ground‑truth labels. In the absence of labels, models cannot optimize the primary task loss directly and must instead depend on self‑generated supervisory signals such as pseudo‑labels or auxiliary self‑supervised tasks such as in Meta Test‑Time Training (MT3) ( \cite{bartler2022mt3}) or entropy‑minimization objectives such as in TENT (\cite{wang2020tent}). These surrogate signals, however, are often noisy and poorly aligned with the original task, leading to error accumulation and training instability (\cite{wang2024continual}). Furthermore, adapting on all generated auxiliary samples imposes significant computational overhead and exacerbates catastrophic forgetting, when low‑confidence or incorrect updates are applied uniformly during test time (\cite{niu2022efficient}).

\begin{figure*}[t]
  \centering
   \includegraphics[width=\textwidth]{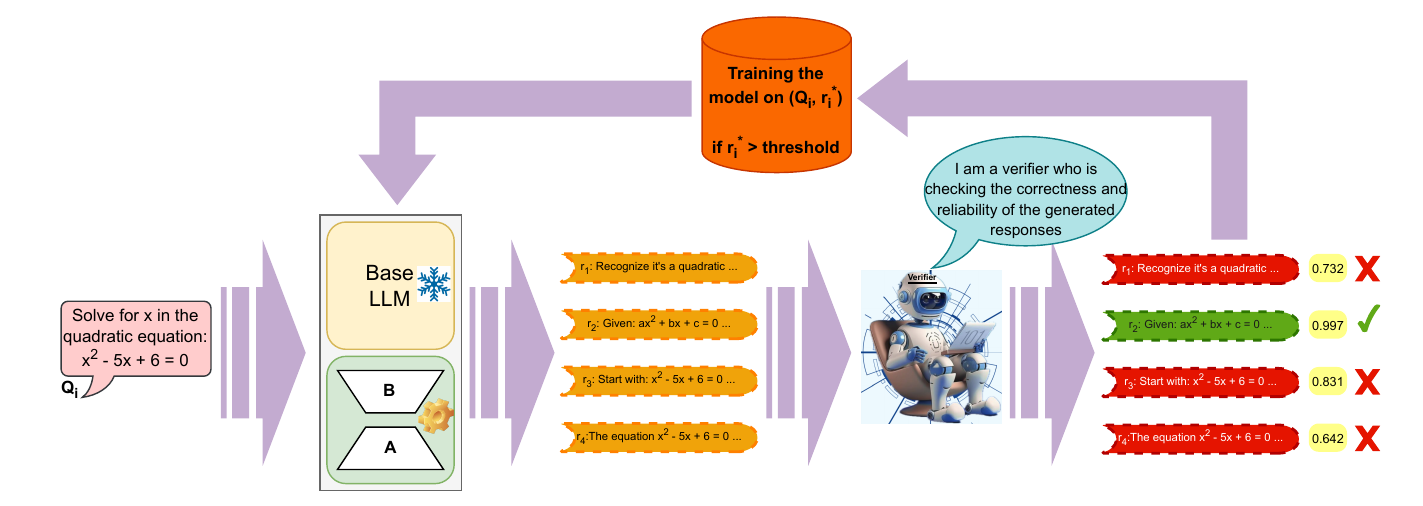}
   \caption{In our VDS‑TTT framework, we proceed through three sequential stages. First, Candidate Generation for Self‑Annotation, where each input question $Q_i$ is passed to the pretrained LLM to produce a set of $N$ candidate responses $\{r_1,...,r_N\}$. Second, Confidence‑Guided Annotation, in which a verifier assigns a reliability score to each $r_j$, and we select the response $r^*$ only if its score exceeds a predefined threshold $\tau$, thereby forming the test‑time training example $(Q_i, r^*)$. Finally, Test‑Time Training, where we fine‑tune the model by optimizing its parameters on the resulting pseudo‑labeled dataset.}
   \label{fig:onecol}
\end{figure*}

In this work, we introduce Verifier‑Driven Sample Selection Test‑Time Training (VDS‑TTT), a new framework that synthesizes high confidence supervisory signals from unlabeled inputs. For each test example, VDS‑TTT first generates a diverse set of candidate solutions by repetitively sampling from a large language model with a positive temperature. A pre-trained domain‑specific verifier (for instance, a suite of unit tests in mathematical reasoning tasks) then scores each candidate on correctness. Test examples with all candidates receiving below acceptable confidence score from the verifier are not considered for use in training. For other test examples, candidates above a confidence threshold are retained as pseudo‑labels to be used in training. In short, the model undergoes preference‑based adaptation using only the verifier‑approved samples (Figure~\ref{fig:onecol}). By leveraging a verifier to filter noisy outputs, VDS‑TTT aligns optimization with the primary task objective, prevents catastrophic forgetting through modification of only LoRA adapter parameters, and self improves by dynamically adjusting to varying domain shifts across continuous test streams. Empirical results demonstrate that VDS‑TTT achieves substantially robust adaptation compared to other approaches addressing label‑scarce, real‑world scenarios.

Our main contributions are as follows:
\begin{itemize}
    \item We introduce VDS‑TTT, a new framework, a first to leverage a learned verifier to select a high confidence pseudo‑label for use in supervised-finetuning at test time, addressing the absence of ground‑truth labels during test time, and then fine‑tunes low‑rank LoRA adapter parameters to self-improve the base LLM on the fly.

    \item Through comprehensive experiments across multiple benchmarks, we demonstrate that VDS‑TTT consistently improves test‑time accuracy compared to baseline without test time training, significantly outperforming methods that simply select samples based on verifier‑computed confidence scores. This validates the effectiveness of integrating pseudo‑label‑based fine‑tuning into the test time pipeline.

    \item We show that increasing test time compute by increasing the number of test‑time training iterations yields  performance improvements, eventually matching or exceeding an oracle verifier at the first iteration, highlighting both the efficiency and robustness of our approach under heterogeneous solution distributions.

    \item Experiments across three diverse benchmarks and three state‑of‑the‑art LLMs demonstrate that VDS‑TTT yields up to a 32.29\% relative improvement over the base model and a 6.66\% gain compared to verifier-based methods without test‑time training.
\end{itemize}

\section{Related Work} \label{sec:Related Work}
\label{related_work}
At test time, ground‑truth labels are inherently unavailable, which prohibits the direct application of standard supervised objectives. Consequently, a rich line of work has emerged in the image and text domains around the development of self-supervised or unsupervised learning signals at the test time.  

In the image domain, numerous self‑supervised tasks have demonstrated effectiveness for this purpose, including rotation prediction (\cite{sun2020test}), meta‑learning adaptations (\cite{bartler2022mt3}), and contrastive learning frameworks (\cite{burns2023weak}). By tailoring these auxiliary losses to the current test distribution, such methods encourage the feature extractor to remain robust under domain shifts. However, applying these techniques in the text domain can alter the input semantics.

In the language modeling literature, different works tried to obtain unknown ground-truth labels to improve the model that can be categorized into extra-memory approaches through retrieval, internal or external feedback derived directly from model predictions, and reinforcement learning (RL) approaches. 

For extra-memory approaches, retrieval-based solutions such as \cite{hardt2023test} show positive gains in terms of perplexity across more than 20 downstream tasks. This is done by retrieving semantically similar contexts and performing a single gradient update on each of a handful of neighbors. Large retrieval indices incur significant storage and computational overhead: the full index can occupy between 810GB and 2.1TB. Moreover, naively selecting the top‑N highest‑scoring neighbors could often retrieve redundant or non‑informative examples. To overcome this memory limitation, \cite{hubotter2024efficiently} proposed a data‑selection algorithm called SIFT that unifies retrieval with active learning. SIFT quantifies the model’s posterior uncertainty about a prompt after fine‑tuning and selects examples with the least uncertainty. However, they still require memory to store information.

Another prominent category leverages internal or external feedback derived directly from model predictions. For example, Tent (\cite{wang2020tent}) treats the prediction entropy as a proxy for uncertainty and fine‑tunes the model during test time by minimizing this entropy. This approach did not consider model collapse into trivial constant output, which needs to be regularized (\cite{press2024entropy}). 

Beyond fully unsupervised signals, certain task‑specific frameworks exploit small amounts of human feedback or external reward functions as high‑quality supervision (\cite{ji2025test}). These reward feedbacks are used in self-training approaches vi RL. ReST (\cite{gulcehre2023reinforced}) and ReST-MCTS (\cite{zhang2024rest}) are the most well-known RL-based approaches. 

ReST \cite{gulcehre2023reinforced} tries to formulate a growing-batch offline RL paradigm for LLMs but suffers from two issues. First is that it requires annotated data as during each Grow step, ReST samples new outputs from the current policy which, together with the original annotated data, are scored and retained only if their reward exceeds an iteratively increasing threshold. Second, the method uses a reduced learning rate across multiple Improve iterations to prevent overfitting, which could be better circumvented using parameter-efficient techniques. 

Building on ReST, ReST-MCTS (\cite{zhang2024rest}) targets reasoning-intensive tasks by embedding a Process Reward Model (PRM) within a Monte Carlo Tree Search framework to collect and value step-by-step reasoning traces. However, this technique has two major limitations. First, MCTS rollouts guided by a PRM usually require high computational cost that may not be easily applicable during test time. Second, the policy model requires high-quality full reasoning traces with known ground truth labels, which limits its direct application to test-time. 

Most recently, \cite{zuo2025ttrl} proposed an RL-based method that incorporates reward feedback during test-time. To derive a pseudo-label, this selects the answer with majority vote over a large ensemble of model generations and uses this answer for RL training. Specifically, Proximal Policy Optimization (PPO) (\cite{schulman2017proximal}) and Group Relative Policy Optimization (GRPO) (\cite{shao2024deepseekmath}) optimize the policy at test-time. PPO follows a traditional actor–critic paradigm, generating one output per prompt, estimating advantages with a learned value network, and updating both policy and value networks with a clipped surrogate objective. In contrast, GRPO dispenses with a separate critic by sampling a large group of responses per prompt, computing group‐normalized rewards to estimate advantages, and updating only the policy with surrogate clipping and a KL penalty. As mentioned in \cite{zuo2025ttrl}, RL-based methods typically depend on the pretrained distribution and are sensitive to the input difficulty level. This can be seen further in our experimental results section.

To alleviate the aforementioned challenges in the literature, we propose VDS‑TTT, a continuous self‑improvement paradigm that applies reward‑based feedback at test-time. Our method has the following advantages: (1) does not require any labeled data; (2) simple and memory‑efficient wihout extra memory requirements and expensive rollouts; (3) simpler to implement, offers greater stability in terms of convergence due to its low-variance gradient estimates \cite{mukobi2023superhf}, and is more computationally efficient compared to existing approaches; (4) we update only low‑rank LoRA adapter parameters rather than the entire model to be better suited to test-time that requires on-the-fly specialization while keeping model's original performance since it is a lightweight adaptation so it can not dramtically change the base model.

\section{Methodology} \label{sec:Methodology}
In this section, we introduce VDS‑TTT (Algorithm \ref{alg:vds-ttt}), our new LLM‑based, fully test‑time training method. VDS‑TTT operates in three sequential stages designed to improve model robustness at test time. First, the pretrained language model generates a diverse set of candidate solutions for each input problem, leveraging its capacity to explore multiple hypotheses. Next, a verifier-based scoring function evaluates each candidate. If all candidates are low scoring (below a pre-defined acceptable threshold), the input problem and its candidate solution(s) are not considered for use in further training. Otherwise, the highest‑scoring output is selected as the provisional solution and paired with the input problem to form the sample data for training. %Finally, by applying a predefined score threshold, we filter these top solutions to construct a curated test‑time training dataset;
The model is then fine‑tuned on this sample data during testing, enabling it to adapt dynamically to distributional shifts and improve its performance on challenging inputs.
%SPM Are we using more than one candidate for training?

\begin{algorithm}[tt]
\caption{Verifier-Driven sample Selection for Test-Time Training (VDS-TTT)}
\label{alg:vds-ttt}
\begin{algorithmic}[1]

  \Require 
    Pretrained LLM $f_{\theta_0}$,  
    verifier score function $s(\cdot,\cdot)$,  
    temperature $T$,  
    number of samples $N$,  
    score threshold $\tau$,  
    LoRA adapter steps $M$,  
    learning rate $\eta$  
  \Ensure  
    Adapted adapter parameters $\Delta$  

  \State Initialize adapter $\Delta \gets \mathbf{0}$  
  \ForAll{test query $q_i$}
    \Comment{Stage 1: Candidate Generation for Self‑Annotation}
    \State $\mathcal{R} \gets \emptyset$
    \For{$j = 1$ to $N$}
      \State Sample response $r_{ij} \sim f_{\theta_0}\bigl(\cdot \mid q_i; T\bigr)$  \Comment{temperature sampling}
      \State Extract final answer $a_{ij}$ from $r_{ij}$
      \State $\mathcal{R} \gets \mathcal{R} \cup \{(r_{ij}, a_{ij})\}$
    \EndFor

    \Comment{Stage 2: Confidence‑Guided High‑Quality Annotation}
    \State Compute scores $s_{ij} \gets s\bigl(r_{ij}, a_{ij}\bigr)$ for each $(r_{ij},a_{ij}) \in \mathcal{R}$
    \State Let $j^* = \arg\max_{1 \le j \le N} s_{ij}$
    \If{$s_{i j^*} < \tau$}
      \State \textbf{continue}  \Comment{skip low‐confidence query}
    \EndIf
    \State Set pseudo‑label $(r_i, a_i) \gets (r_{i j^*}, a_{i j^*})$
    \State Record score $s_i \gets s_{i j^*}$

    \Comment{Stage 3: Test‑Time Training (LoRA Adaptation)}
    \For{$m = 1$ to $M$}
      \State Compute loss  
      \[
        \mathcal{L}(\Delta) 
        = -\sum_{t=1}^{|r_i|} \log\,f_{\theta_0+\Delta}\bigl(r_{i,t}\mid q_i, r_{i,<t}\bigr)
      \]
      \State $\Delta \gets \Delta - \eta \,\nabla_{\Delta}\,\mathcal{L}(\Delta)$
    \EndFor

  \EndFor

  \State \Return $\Delta$

\end{algorithmic}
\end{algorithm}

\subsection{Candidate Generation for Self‑Annotation} \label{sec:Candidate Generation for Self‑Annotation}
In our approach, we automatically construct a labeled dataset $D = \{q_i, r_{ij}, a_{ij}\}_{j=1}^N$ without relying on human annotations, where $q_i$, $r_{ij}$, $a_{ij}$, and $N$ respectively denote input question, LLM generated detailed step by step responses, LLM generated corresponding final answer, and  number of generated responses per input question. We do this by invoking the pretrained language model to generate for each question a diverse set of candidate answers, along with detailed step by step responses by employing temperature sampling, in which the softmax temperature $T$ is set at test time to control the randomness of the model’s output, thus letting the model explore more unconventional and less likely outputs (\cite{renze2024effect}). 

\subsection{Confidence-guided High-quality Data Annotation} \label{sec:Confidence-guided High-quality data annotation}
At this stage, for each generated candidate response, we must determine which ones are suitable for inclusion in model training. Two paradigms exist: Verifier‐Free methods (VF), which directly distill successful search or “thinking” traces without any external check, and Verifier‐Based (VB) methods, which interleave a verification step to guide the search process. As demonstrated by \cite{setlur2025scaling}, VF methods exhibit poor scalability when the base pretrained LLM produces a heterogeneous distribution over correct solution traces and is thus suboptimal, whereas VB methods maintain robust performance under these conditions. Consequently, we adopt a VB selection strategy to ensure both scalability and accuracy in assembling our test time training data.

For the $N$ candidate responses we apply a Best‑of‑N selection strategy: our pretrained verifier assigns a real‑valued score to each generated response, and we choose the response $r^* = argmax_{y \in \{r_1,...,r_N\}} s(r)$  with the highest score as the provisional answer. To ensure both accuracy and informativeness, we further impose a score threshold $\tau$: any response whose verifier score $s(r)$ falls below $\tau$ is discarded, so that only those candidates deemed sufficiently confident and high‑quality are used to train the model, thus aligning better with the query task,. This two‑step process, first selecting the single highest‑scoring response and then filtering by a minimum confidence threshold, guarantees that our generated training data comprises diverse yet reliably correct solution traces.

\subsection{Test Time Training} \label{sec:Test Time Training}
For each test query $q_i$ that is selected for TTT, we construct a test‑time training set $D_{TTT} = \{q_i, r_i, a_i, s_i\}$ where $s_i$ is the corresponding score provided by the Best‑of‑N verifier procedure. By treating each unlabeled response as labeled according to its confidence score, this method efficiently leverages unlabeled data without human annotation. Then, we fine‑tune the pretrained LLM by minimizing the empirical risk: 
\begin{equation}
    min_{\Delta} \sum_{(q, r, a, s) \in D_{TTT}} L(LM_{\Delta}(q, r), \hat{y})
\end{equation}
where $\Delta$ denotes the LoRA \cite{hu2022lora}\ adapter parameters while the original model weights remain frozen. Freezing the base LLM and updating only LoRA adapter parameters preserves model stability, and avoids catastrophic forgetting while ensuring rapid convergence during test‑time training. By leveraging pseudo‑labels in a one‑sample self‑supervised manner, we enable on‑the‑fly model adjustment with each high confidence test instance, improving robustness under distributional shifts and enabling continuous self improvement.

\section{Experiments}
\subsection{Settings} \label{sec:Settings}
\textbf{Datasets. } In our experimental evaluation, we assess VDS‑TTT on three established mathematical reasoning benchmarks: GSM8K, Math‑500, and AIME1983‑2024. GSM8K \cite{cobbe2021training} comprises 1319 grade‑school math word problems, segmented test instance, each requiring 2–8 sequential elementary arithmetic operations to arrive at a final numeric answer. Math‑500 is a curated subset of 500 problems drawn from the MATH dataset (12.5K competition‑level questions with full, step‑by‑step solutions), \cite{hendrycks2021measuring}, covering seven domains—Prealgebra, Algebra, Number Theory, Counting and Probability, Geometry, Intermediate Algebra, and Precalculus—as selected in \cite{lightman2023let}. Finally, AIME1983‑2024  comprises challenging problems from the American Invitational Mathematics Examination (AIME), collected over four decades (1983–2024). These competition-level problems require advanced reasoning, often combining algebraic manipulation with creative problem-solving and geometric insight. The American Mathematics Competitions (AMC) dataset contains 83 samples comprising a series of progressively challenging mathematical tests designed for middle and high school students. We report exact‑match accuracy on held‑out test splits, extracting final answers via regular‑expression parsing and verifying equivalence with ground truth. 

\textbf{Baseline Models. } To validate the generality of VDS‑TTT, we apply our framework to four state‑of‑the‑art LLMs: Llama‑3.2‑1B‑Instruct, DeepSeek‑R1‑Distill‑Qwen‑1.5B, Llama‑3.2‑3B‑Instruct, and LLaMA-3.1-8B-Instruct. Moreover, to reduce computational overhead, we adopt a lightweight reward model, Skywork-o1-Open-PRM-Qwen-2.5-1.5B.

\subsection{Implementation details} \label{sec:Implementation details}
In our experiments, we systematically vary the number of candidate responses $N \in \{2,4,8,16\}$ to evaluate its impact on Best‑of‑N selection performance. We adopt a stringent verifier confidence threshold $\tau = 0.99$ for GSM8K and Math‑500—mirroring pseudo‑labeling best practices that apply high thresholds to filter out noisy labels, but relax $\tau$ to 0.9 for the more challenging AIME1983‑2024 to prevent discarding the vast majority of low‑scoring samples if $\tau = 0.99$. Furthermore, we empirically concluded that updating only low‑rank LoRA adapter parameters substantially outperforms full fine‑tuning of deeper or randomly selected layers. Consequently, we integrate LoRA modules into all key projection components—q\_proj, k\_proj, v\_proj, o\_proj—as well as the MLP sublayers (mlp\_gate\_proj, mlp\_up\_proj, mlp\_down\_proj) of our base LLM. We set the LoRA rank to 128 in most cases where sufficient data is available for test-time training. For low-resource scenarios, such as AIME 2024, we reduce the LoRA rank to 8 to prevent overfitting and maintain training stability. 

All experiments were performed on a machine equipped with an NVIDIA Tesla V100 PCIe GPU with 32GB of memory. As a representative example of total VDS-TTT execution time, applying the LLaMA-3.2-1B-Instruct model to the Math-500 dataset with 2, 4, 8, and 16 candidate responses resulted in total runtimes of approximately 1h30m, 2h10m, 3h6m, and 4h30m, respectively.

\subsection{Results and Discussion} \label{sec:Results and Discussion}
We first assess how VDS‑TTT enhances model accuracy across different sampling budgets. Table \ref{table:1} reports exact‑match accuracy for our three baselines—Llama‑3.2‑1B‑Instruct, Llama‑3.2‑3B‑Instruct, and DeepSeek‑R1‑Distill‑Qwen‑1.5B—on GSM8K, Math‑500, and AIME1983–2024 using $N=2, 4, 8, 16$ candidate responses. For each $N$, we compare three settings: (1) Base: the frozen model without any Test Time Training; (2) VB (Verifier-Based): where we generate $N$ samples and select the single highest‑scoring response via our verifier; and (3) VDS‑TTT: which fine‑tunes LoRA adapters on all verifier‑selected, high‑confidence pseudo‑labels. 

The most dramatic improvements occur on AIME1983–2024, where Qwen‑1.5B’s Base accuracy is near zero (0.54\% for N=2), VB selection raises it to 0.88\%, and just a single iteration of our VDS‑TTT boosts it to 4.22\%. With N=16, VDS‑TTT accuracy increases to 6.96\%. These results  highlight VDS‑TTT’s ability to extract reliable pseudo‑labels and adapt models on‐the‐fly, even when the pretrained base model lacks prior competence in the test domain. Moreover, the gain between $N=2$ (2–6\% over VB) and $N=4$  (1–5\%) exceeds that between $N=8$ (1–3\%) and $N=16$ (1–2\%), indicating diminishing returns at larger sample sizes and suggesting that $N=4$ often suffices for robust adaptation. Overall, VDS‑TTT not only amplifies verifier‑based selection but also delivers robust, parameter‑efficient adaptation under severe distribution shifts, with the greatest relative gains at smaller candidate budgets.

\begin{table}
  \caption{Accuracy improvements across Math-500, GSM-8K, and AIME1983-2024 benchmarks for Llama3-1B, Qwen-1.5B, and Llama3-3B models using Base, Verifier-Based (VB), and VDS-TTT methods with varying numbers of samples ($N = 2, 4, 8, 16$). The results demonstrate the effectiveness of VDS-TTT in enhancing model performance, particularly in low-resource scenarios.\\}
  \label{table:1}
  \centering
  \setlength{\tabcolsep}{3.5pt}
  \begin{tabular}{lcccccccccc}
    \toprule
    \multirow{2}{*}{Model Name} & \multirow[c]{2}{*}{N} & \multicolumn{3}{c}{Math-500} & \multicolumn{3}{c}{GSM-8K} & \multicolumn{3}{c}{AIME1983-2024} \\
    \cmidrule(r){3-11}
     &  & Base & VB & VDS-TTT & Base & VB & VDS-TTT & Base & VB & VDS-TTT \\
    \midrule
    \multirow{4}{*}{Llama3-1B} & 2  & \multirow{4}{*}{20.8} & 24.00 & 28.00 & \multirow{4}{*}{40.18} & 52.31 & 55.88 & \multirow{4}{*}{3.42} & 6.22 & 8.37 \\
                              & 4  &                        & 26.80 & 30.60 &                        & 59.67 & 62.40 &                        & 6.43 & 9.29 \\
                              & 8  &                        & 33.60 & 36.60 &                        & 63.08 & 63.84 &                        & 9.75 & 12.35 \\
                              & 16 &                        & 36.80 & 37.20 &                        & 71.52 & 72.47 &                        & 11.36 & 12.63 \\
    \midrule
    \multirow{4}{*}{Qwen-1.5B} & 2  & \multirow{4}{*}{19.20} & 24.80 & 26.60 & \multirow{4}{*}{21.15} & 24.46 & 27.56 & \multirow{4}{*}{0.54} & 0.88 & 4.22 \\
                               & 4  &                        & 28.60 & 29.80 &                        & 25.93 & 30.09 &                        & 1.29 & 4.31 \\
                               & 8  &                        & 33.40 & 34.20 &                        & 26.93 & 28.13 &                        & 1.93 & 5.11 \\
                               & 16 &                        & 35.00 & 36.60 &                        & 34.27 & 35.12 &                        & 4.89 & 6.96 \\
    \midrule
    \multirow{4}{*}{Llama3-3B} & 2  & \multirow{4}{*}{31.80} & 37.00 & 40.20 & \multirow{4}{*}{73.09} & 80.29 & 81.11 & \multirow{4}{*}{14.47} & 18.01 & 24.67 \\
                               & 4  &                        & 39.40 & 41.60 &                        & 84.15 & 85.33 &                        & 21.44 & 25.51 \\
                               & 8  &                        & 42.60 & 44.80 &                        & 85.14 & 85.78 &                        & 21.65 & 26.16 \\
                               & 16 &                        & 44.00 & 46.40 &                        & 85.90 & 88.44 &                        & 22.94 & 27.59 \\
    \bottomrule
  \end{tabular}
\end{table}

Table \ref{table:2} presents a comparison between our proposed VDS-TTT method and the very recently published TTRL method that uses RL for TTT \cite{zuo2025ttrl}. Results show that VDS-TTT outperforms TTRL on the instruction-tuned LLaMA-3.1-8B-Instruct model. This result is particularly striking given that, as acknowledged by TTRL’s authors, reinforcement-learning-based strategies can suffer from sensitivity to query difficulty level, heavy reliance on strong pretraining on similar distributions, and potential training collapse under some conditions. As shown on the challenging AIME2024 dataset with LLaMA-3.1-8B-Instruct, where the model is not specifically tuned for math, TTRL has no effect during test time, while VDS-TTT achieves a 6.7\% improvement. Taken together, these findings underscore the practical advantages of VDS-TTT: it achieves competitive and even superior test time performance, is straightforward to implement, and entails significantly less computational overhead compared to the RL-based approaches.

%Table \ref{table:2} presents a comparison between our proposed VDS-TTT method and the very recently published TTRL method that uses RL for TTT \cite{zuo2025ttrl}. While TTRL outperforms VDS-TTT on the two base models (Qwen2.5-Math-1.5B and Qwen2.5-Math-7B), VDS-TTT decisively outshines TTRL on the instruction-tuned LLaMA-3.1-8B-Instruct model. This result is particularly striking given that, as acknowledged by TTRL’s authors, reinforcement-learning-based strategies can suffer from sensitivity to query difficulty level, heavy reliance on strong pretraining on similar distributions, and potential training collapse under some conditions. As shown on the challenging AIME2024 dataset with LLaMA-3.1-8B-Instruct, where the model is not specifically tuned for math, TTRL has no effect during test time, while VDS-TTT achieves a 6.7\% improvement. Taken together, these findings underscore the practical advantages of VDS-TTT: it achieves competitive and even superior inference time performance, is straightforward to implement, and demands significantly less computational overhead.

\begin{figure}[ht]
    \centering
    \begin{subfigure}{0.3\textwidth}
        \includegraphics[width=\linewidth, height=2.5cm]{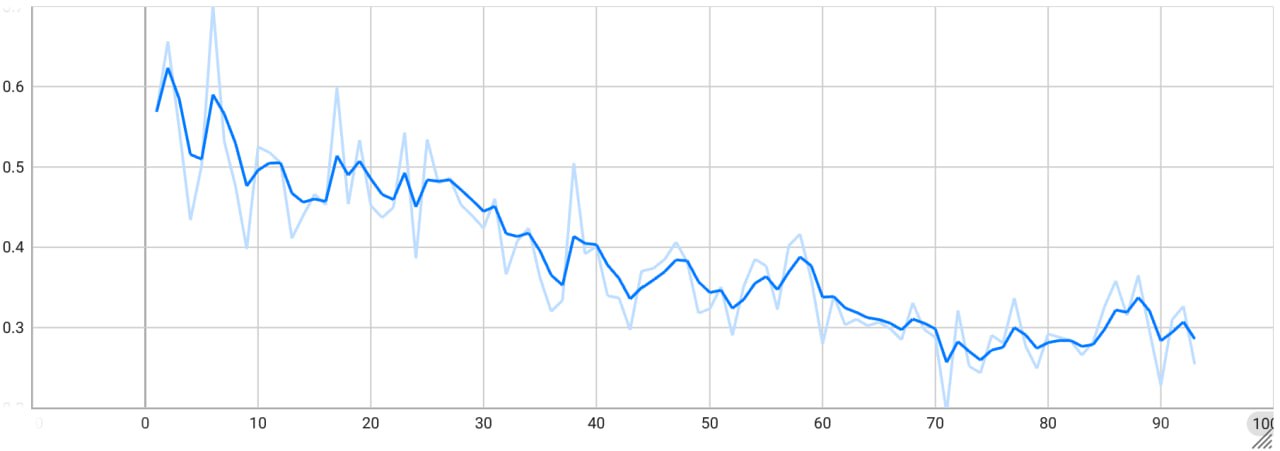}
        \caption{Llama-3.2-1B-Instruct on Math-500 for $N=4$}
        \label{fig:1a}
    \end{subfigure}
    \hfill
    \begin{subfigure}{0.3\textwidth}
        \includegraphics[width=\linewidth, height=2.5cm]{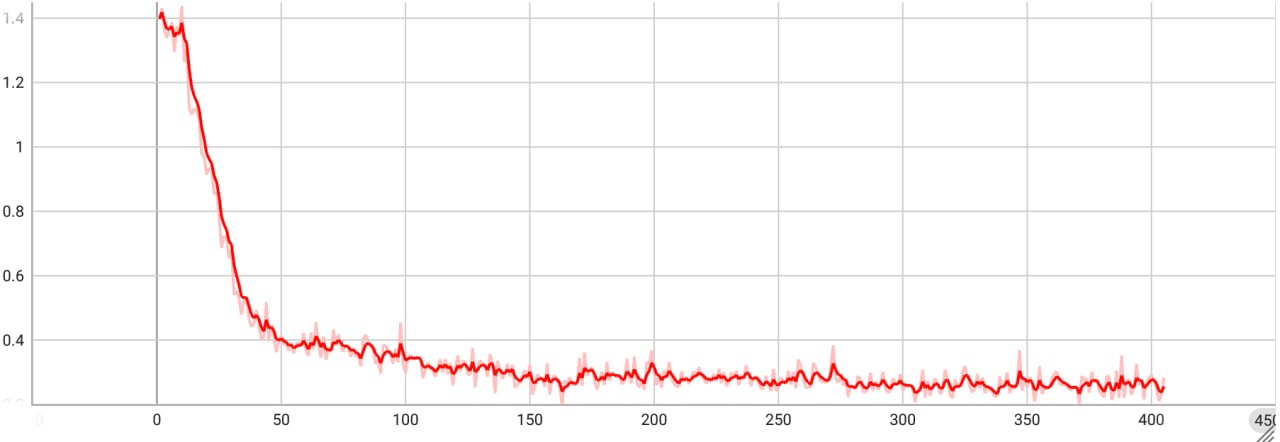}
        \caption{Llama-3.2-3B-Instruct on GSM-8K for $N=2$}
        \label{fig:1b}
    \end{subfigure}
    \hfill
    \begin{subfigure}{0.3\textwidth}
        \includegraphics[width=\linewidth, height=2.5cm]{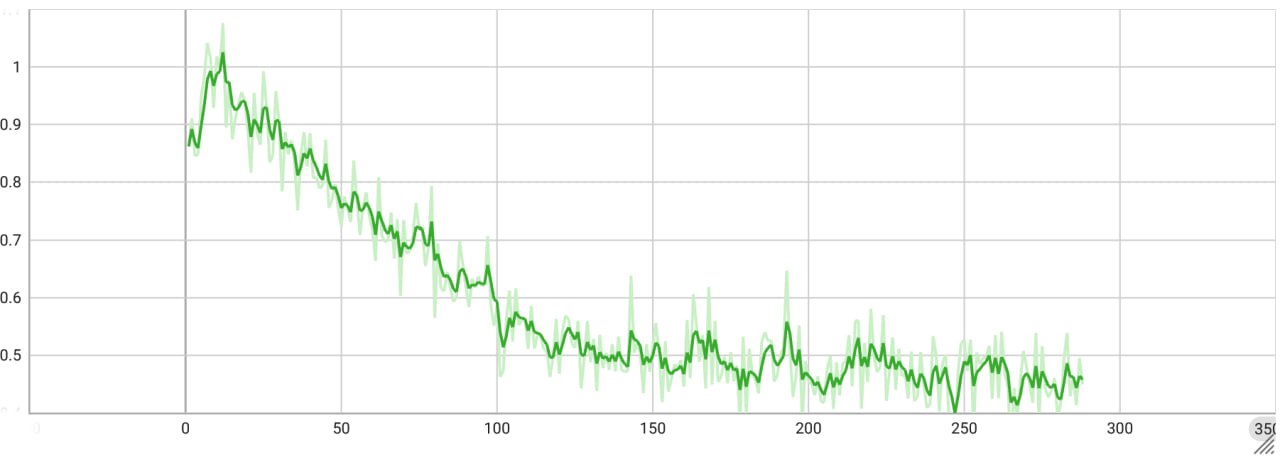}
        \caption{DeepSeek-R1-Distill-Qwen-1.5B on AIME for $N=8$}
        \label{fig:1c}
    \end{subfigure}
    \caption{Three instances of TTT loss curves}
    \label{fig:1}
\end{figure}

Figure \ref{fig:1} plots the test‑time training loss incurred by VDS‑TTT for three representative model–benchmark–sampling configurations. In all cases, the loss curves exhibit a smooth, monotonic decrease and plateau at low values, confirming that VDS‑TTT consistently adapts model parameters to the test‑time distribution and achieves reliable convergence under diverse settings.

\begin{figure}[ht]
    \centering
    \begin{subfigure}{0.45\textwidth}
        \includegraphics[width=\linewidth, height=4cm]{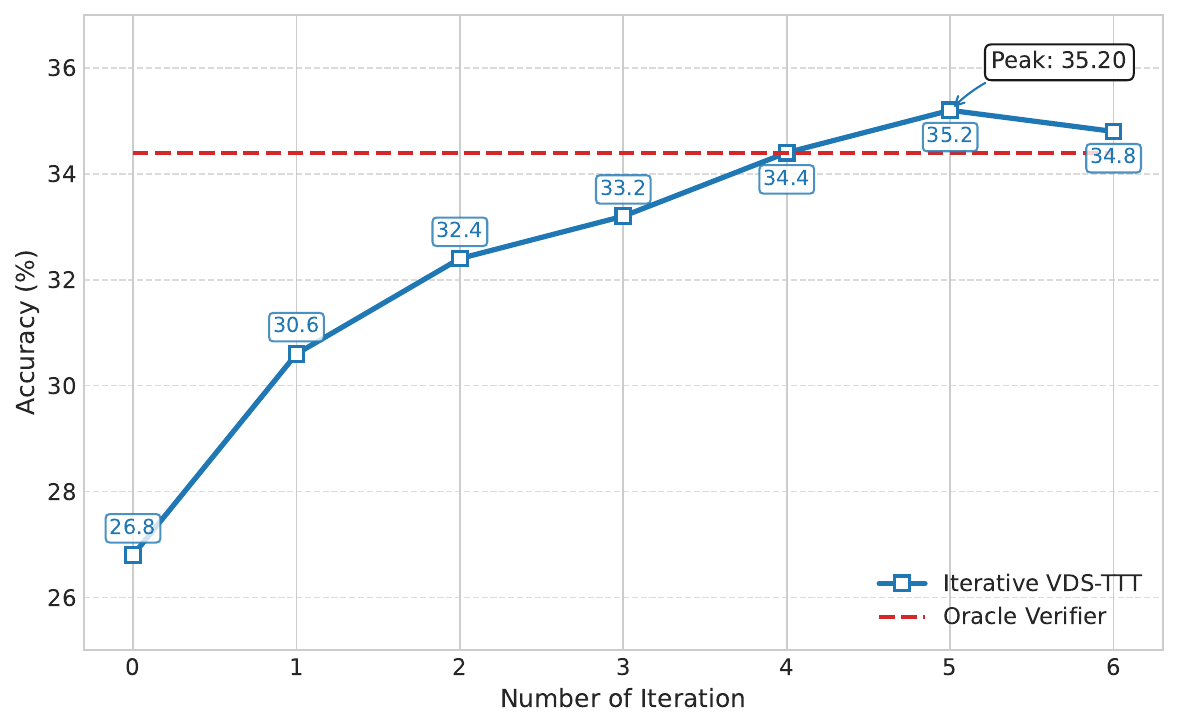}
        \caption{Llama-3.2-1B-Instruct on Math-500 for $N=4$}
        \label{fig:2a}
    \end{subfigure}
    \hfill
    \begin{subfigure}{0.45\textwidth}
        \includegraphics[width=\linewidth, height=4cm]{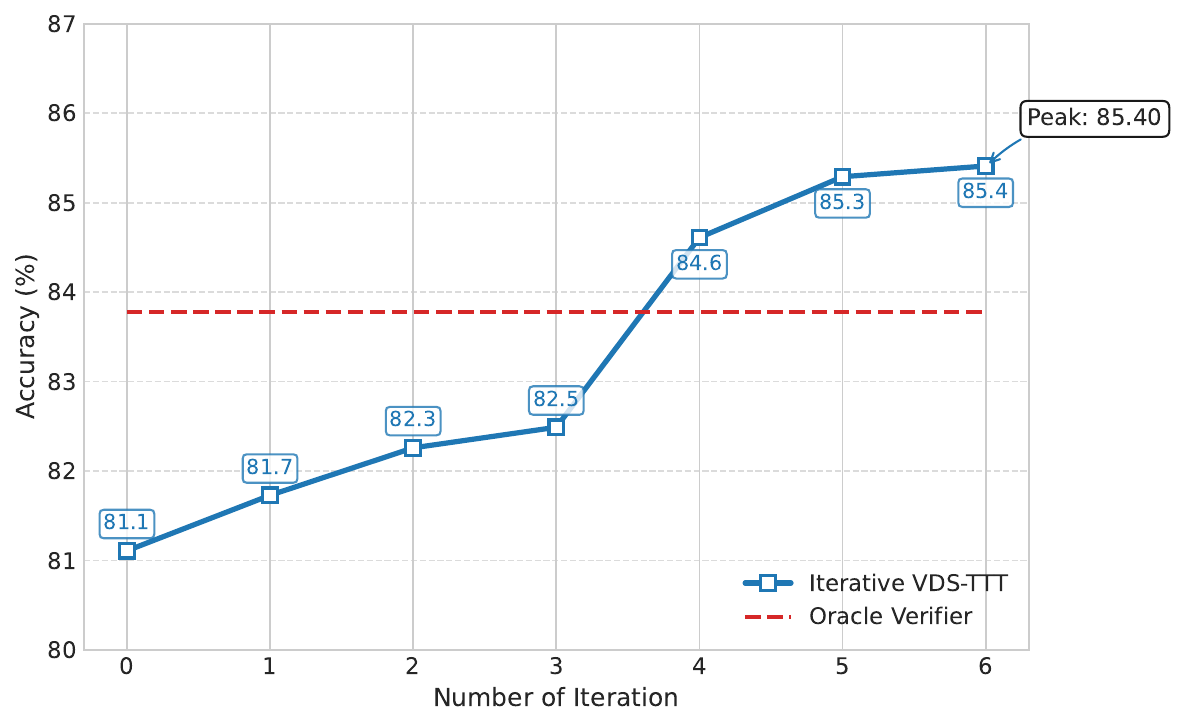}
        \caption{Llama-3.2-3B-Instruct on GSM-8K for $N=2$}
        \label{fig:2b}
    \end{subfigure}
    \hfill
    \caption{Iterative VDS-TTT results}
    \label{fig:2}
\end{figure}

Figure \ref{fig:2} illustrates the performance trajectory of the iterative VDS-TTT strategy, where the VDS-TTT procedure is applied multiple iterations to the same model. In this figure, the x-axis denotes the number of iterations (i.e., how many times VDS-TTT has been applied), while the y-axis indicates the corresponding accuracy at each iteration. This iterative process reveals a clear trend: model performance consistently improves with successive iterations—up to a certain point. As shown in Table \ref{table:1} (AIME task), even a single round of VDS-TTT (iteration 1) enables the model to acquire task-specific capabilities, despite having no prior exposure. This motivated us to investigate whether further iterations could yield additional gains. For example, one iteration of VDS-TTT achieves an accuracy of 81.11\%, and a second iteration—starting from the fine-tuned checkpoint—leads to further improvements, as the model becomes increasingly capable of generating higher-quality task-relevant outputs. However, due to inherent model capacity limitations, this iterative improvement eventually exhibits diminishing returns. As shown in Figure \ref{fig:2}, performance plateaus beyond a certain point, additional iterations result in minor fluctuations rather than continued gains. 

The red dashed line in Figure \ref{fig:2} represents the accuracy achieved by the Oracle Verifier. An Oracle Verifier is a theoretical tool that assumes access to ground-truth answers, allowing it to perfectly assess the correctness of model-generated responses. It serves as an upper-bound benchmark to evaluate and compare the effectiveness of practical verification methods used during test time. If an oracle verifier is employed at the first iteration, the model accuracy corresponds to the red dashed line. Remarkably, in several iterations, iterative VDS‑TTT surpasses an Oracle verifier demonstrating that self‑improvement through multiple iterations can exceed even the ideal “best‑of‑N” selection at iteration zero. This finding highlights VDS‑TTT’s unique ability to not only bootstrap performance from scratch but also to progressively refine model proficiency beyond static verification alone.

A critical insight from Figure \ref{fig:2} is determining the optimal early stopping point for the iterative process. To address this, we define a maximum number of iterations, $T_{\text{max}}$, based on computational budget constraints. We then identify the optimal iteration $t^*$, which yields the highest achievable accuracy, using the following criterion: if the difference in accuracy between two consecutive iterations falls below a threshold $\epsilon$ for two successive steps, the process should be halted. Formally, if $A^{(t)}$ denotes the accuracy at iteration $t$, we stop when the accuracy gains between iterations become marginal, indicating convergence. This approach ensures an efficient balance between performance and computational cost by avoiding unnecessary iterations beyond $t^*$.

\begin{equation}
    t^* = \min\{t=2,...,T_{max}: |A^{(t)} - A^{(t-1)}| < \epsilon \text{ and } |A^{(t-1)} - A^{(t-2)}| < \epsilon\}
\end{equation}

%performance consistently improves with successive iteration, up to a certain point since when we apply the VDS-TTT for the first time (Iteration 1) next time (Iteration 2) the model is able to produce higher quality responses which gives rise to more number of true responses and highher accuracy.

\begin{table}
  \caption{Comparison of our proposed framework, VDS-TTT, with the TTRL (\cite{zuo2025ttrl}) approach across three challenging benchmarks. * indicates results from GRPO paper (\cite{hu2025open}) \\}
  \label{table:2}
  \centering
  \setlength{\tabcolsep}{3.5pt}
  \begin{tabular}{llcccc}
    \toprule
    Model Name & Methods & AIME 2024 & AMC & Math-500 & Average \\    
    \midrule
    \multirow{6}{*}{LLaMA-3.1-8B-Instruct} & Base model*                        & 3.3   & 19.3  & 47.8  & 23.5  \\
                                            \cmidrule(r){2-6}
                                            & TTRL                              & 3.3   & 32.5  & \textbf{61.8}  & 32.5  \\
                                            & VDS-TTT                           & \textbf{10.0}  & \textbf{38.5}  & 54.2  & \textbf{38.3}  \\
                                            \cmidrule(r){2-6}
                                            & $\Delta$\textsubscript{TTRL}      & 0.0   & +13.2 & \textbf{+14.0} & +9.0 \\
                                            & $\Delta$\textsubscript{VDS-TTT}   & \textbf{+6.7}  & \textbf{+19.2} & +6.4  & \textbf{+10.8}\\
    \bottomrule
  \end{tabular}
\end{table}

\section{Conclusion} \label{sec:conclusion}
We have presented Verifier‑Driven sample Selection Test‑Time Training (VDS‑TTT), a simple yet powerful framework for self-supervised continuous improvement of an LLM model. For each test question, that the LLM responds with high confidence, we choose the single best answer as per a learned verifier to train/adapt the base LLM model on a continuing basis. By using just one response, the test time compute addition is kept low, but the gain in performance is significant, as can be seen from experimental comparisons  over other recent methods.  
%adapting pretrained language models to unlabeled, out‑of‑distribution data by leveraging a learned verifier to curate high‑quality pseudo‑labels at inference time. 
By generating 
$N$ candidate responses per query, scoring them with a verifier, and filtering via a confidence threshold, VDS‑TTT constructs a reliable test‑time training set on which only low‑rank LoRA adapters are updated. This parameter‑efficient adaptation mechanism enables rapid convergence and preserves the stability of the base model. Our results confirm that verifier‑guided pseudo‑labeling addresses the absence of ground‑truth labels very well, even under structurally novel reasoning tasks.

\emph{Limitations:}
While VDS‑TTT delivers strong adaptation performance on mathematical reasoning benchmarks, its reliance on a verifier trained exclusively on math problems constrains its applicability to other domains. In settings where the underlying verifier has not been exposed to domain‑specific reasoning patterns, such as code generation, commonsense QA, or multi‑modal tasks, the pseudo‑labels it produces may be unreliable, limiting the effectiveness of test‑time training. Addressing this limitation will likely require either (1) a general‑purpose verifier capable of assessing candidate responses across diverse tasks, or (2) a mixture‑of‑experts verifier architecture that dynamically activates specialized verification modules based on the input prompt. Such a design would preserve VDS‑TTT’s parameter efficiency and test time adaptability while broadening its scope beyond mathematics, thus inviting further exploration into expert routing mechanisms and verifier calibration. Looking forward, we plan to suitably extend VDS‑TTT. 

\emph{Broader Impact:} This paper introduces a framework for actively training and adapting a model during test time. By updating on incoming test examples, our approach dynamically aligns the model with new or evolving data distributions in real time and substantially enhances robustness under unpredictable conditions. On the downside, this continual adaptation incurs additional computational overhead, which may pose challenges for large-scale or resource‑constrained deployments.

\medskip

{
\small

\bibliographystyle{plainnat} % or 'abbrvnat' if preferred
\bibliography{references}    % assumes references.bib

\end{document}

